%% file: main_camera.tex
\documentclass[sigconf]{acmart}
\AtBeginDocument{%
  \providecommand\BibTeX{{%
    Bib\TeX}}}

\usepackage{xspace}
\acmConference[DAC'24]{Design Automation Conference}{June 23--27,
  2024}{San Francisco, CA}
\newcommand{\method}{{\textsc{Lotus}}\xspace}
\usepackage{footmisc}
\usepackage[inline]{enumitem}
\usepackage{textcomp}
\usepackage{subfig}  % package for subfloat

\usepackage{booktabs}
\usepackage{multirow}
\usepackage{enumitem}% http://ctan.org/pkg/enumitem

\usepackage{cleveref}

\copyrightyear{2024} 
\acmYear{2024} 
\setcopyright{acmlicensed}\acmConference[DAC '24]{61st ACM/IEEE Design Automation Conference}{June 23--27, 2024}{San Francisco, CA, USA}
\acmBooktitle{61st ACM/IEEE Design Automation Conference (DAC '24), June 23--27, 2024, San Francisco, CA, USA}
\acmDOI{10.1145/3649329.3657310}
\acmISBN{979-8-4007-0601-1/24/06}
\begin{document}

\title{\method: learning-based online thermal and latency variation management for two-stage detectors on edge devices}

% \author{
% Yifan Gong$^{*1}$, Yushu Wu$^{*1}$, Zheng Zhan$^{*1}$, Pu Zhao$^{1}$, Liangkai Liu$^{2}$, Chao Wu$^{1}$,  Xulong Tang$^{3}$, Yanzhi Wang$^{1}$ 
% }
\author{
Yifan Gong$^{*1}$, Yushu Wu$^{*1}$, Zheng Zhan$^{*1}$, Pu Zhao$^{1}$, Liangkai Liu$^{2}$, Chao Wu$^{1}$,  Xulong Tang$^{3}$, Yanzhi Wang$^{1}$ 
}
\affiliation{\institution{$^*$Equal Contribution, $^1$Northeastern University, $^2$University of Michigan, $^3$University of Pittsburgh} \country{}}

\input{marcos}
\begin{abstract}
  Two-stage object detectors exhibit high accuracy and precise localization, especially for identifying small objects that are favorable for various edge applications.
  However, the high computation costs associated with two-stage detection methods cause more severe thermal issues on edge devices, incurring dynamic runtime frequency change and thus large inference latency variations.
  Furthermore,  the dynamic number of proposals in different frames leads to various computations over time, resulting in further latency variations. The significant latency variations of detectors on edge devices can harm user experience and waste hardware resources.
  To avoid thermal throttling and provide stable inference speed, we propose \method, a novel framework that is tailored for two-stage detectors to dynamically scale CPU and GPU frequencies jointly in an online manner based on deep reinforcement learning (DRL).
  To demonstrate the effectiveness of \method, we implement it on NVIDIA Jetson Orin Nano and Mi 11 Lite mobile platforms.
  The results indicate that \method can consistently and significantly reduce latency variation, achieve faster inference, and maintain lower CPU and GPU temperatures under various settings. Our code is available at \textcolor{blue}{\url{https://github.com/wuyushuwys/LOTUS.git}}.   
    % \footnote{The source code will be released upon acceptance.}
   % such as search and rescue, tracking, and environmental monitoring by drones.
\end{abstract}

\maketitle



\input{tex/1_intro}

\input{tex/2_related_work}

\input{tex/3_motivation}

\input{tex/4_method}

\input{tex/5_experiments}

\section*{Acknowledgement}
This work is partly supported by the Army Research Office/Laboratory via grant W911-NF-20-1-0167 to Northeastern University, National Science Foundation CCF-1937500, and CNS-1909172.

\footnotesize
\bibliographystyle{ACM-Reference-Format-num}
\bibliography{acmart}


\end{document}

%% file: marcos.tex
\newcommand{\todo}[1]{\textcolor{red}{\sf\bfseries Todo: #1}}
\newcommand{\bred}[1]{\textcolor{red}{\sf\bfseries #1}}
\newcommand{\red}[1]{\textcolor{red}{#1}}
\newcommand{\blue}[1]{\textcolor{blue}{#1}}
\newcommand{\yellow}[1]{\textcolor{yellow}{#1}}
\newcommand{\purple}[1]{\textcolor{purple}{#1}}
\newcommand{\brown}[1]{\textcolor{brown}{#1}}
\newcommand{\cross}[1]{\textcolor{red}{\sout{#1}}}

%% file: tex/1_intro.tex
\section{Introduction}
With the breakthrough of deep neural networks (DNNs), the object detection task has gained rapid development and wide utilization. There is a growing need to run detection models on widespread edge devices for tasks like autonomous vehicles, drone-based environmental monitoring, or mobile inventory management in retail.  Among the detection models, two-stage detectors offer improved detection by refining region proposals and enhancing object classification with robustness in detecting objects of different sizes, which is especially helpful in scenarios where high precision is required. Though enjoying these advantages, they are typically more computation intensive,  consuming high power during on-device inference, leading to rapidly decreasing battery and increasing temperature. Managing the thermal and latency for two-stage detectors on edge devices is thus a challenging yet urgent problem. 

% Deep  neural networks (DNNs)  can achieve superior performance  for various tasks compared with traditional methods.  With the rapid increasing popularity of  mobile phones, it is necessary to run DNN models on mobile phones. However, as DNNs are usually memory and computationally intensive, it typically consumes  high power during on-device inference, leading to rapid decreasing battery and increasing temperature.   Due to the nature of mobile devices whose users are sensitive to battery consumption and device temperature, it is  essential to investigate energy-efficient operations in
% mobile processors, targeting to manage the temperature and power
% consumption below predefined thresholds. 

% However, as DNNs are usually memory and computation-intensive, it typically consumes high power during on-device inference, leading to rapid decreasing battery and increasing temperature.  To manage the power consumption and heat generation,  DVFS (Dynamic Voltage and Frequency Scaling) is proposed to adjust  CPU or GPU voltage-frequency levels at runtime. Furthermore, as two-stage detection models  Although there are many DVFS techniques developed for edge processors, it is still challenging to implement a  suitable DVFS  ideally for runing  DNNs on mobile devices. 

To reduce heat generation and power consumption,  Dynamic Voltage and Frequency Scaling (DVFS) is proposed by adjusting CPU or GPU voltage-frequency (VF) levels at runtime. However, conventional DVFS is designed for the operating system kernel and is thus application-agnostic. Furthermore, CPU and GPU have separate VF scaling algorithms, %causing 
incurring inefficiency of resource utilization within a limited power budget. Besides, although DVFS can reduce power consumption \cite{wu2023moc}, it does not guarantee to solve the overheating problem \cite{sekar2013power} on edge devices. For edge devices such as mobile phones without active cooling methods such as fan control, if the device temperature goes above a threshold, thermal throttling will be activated to decrease the frequency to a very low level for reducing the temperature. Though the latest work zTT \cite{kim2021ztt} manages to address the thermal throttling problem with a joint VF scaling for CPU and GPU, its direct application for two-stage detectors fails to perform well due to the ineffective design %and neglect of 
without incorporating the characteristics of two-stage detection models.

To overcome the limitations of prior works, we propose \method, a learning-based online thermal and latency variation management framework tailored for two-stage detectors on edge devices. To improve user experience, our objective is to reduce the latency variation and keep the temperature as low as possible through DVFS. We start by formulating the problem mathematically, then conduct in-depth analysis and profiling of two-stage detection models to obtain their characteristics.  It turns out that besides the influence of frequency, the dynamic proposal numbers obtained by the Region Proposal Network (RPN) varying across different images also cause significant latency variations. It makes our problem more challenging as the model latency is affected by both external and internal factors. To tackle this issue, we leverage the DRL approach that can handle dynamic and complex environments with varying images and CPU/GPU temperatures. The DRL agent, powered by a single deep Q-network working at different model widths, provides two chances for frequency scaling during each image frame inference.  
%at  different %two detector 
%stages. 
To train the Q-network effectively, \method keeps two separate experience replay buffers. Furthermore, an additional $\epsilon_t$-greedy cool-down action selection is introduced to avoid thermal issues at early DRL training while allowing the agent to learn how to handle high-temperature cases gradually. To show the effectiveness of \method, we implement it on  NVIDIA Jetson Orin Nano and Mi 11 Lite, and the results demonstrate that \method can achieve better performance for both static and dynamic environment settings.  We summarize our contributions as follows.
% To improve user experience, our objective is to reduce the variation of the inference time and keep the temperature as low as possible for the detector through DVFS. This is not an easy task since changing the  frequency with DVFS  will definitely  change the inference latency  of the computationally intensive DNN model, leading to large inference runtime variation. Besides the influence of frequency, we find that the inference latency is affected by other factors inside the model. As the mask-RCNN adopts a two stage architecture,   it has a Region Proposal Network (RPN) in the model to create proposals.  We find that the inference latency is proportional to the number of proposals from the RPN. If the RPN creates more proposals, the inference latency also becomes larger. 
% Thus,  it is challenging to use DVFS to keep the temperature as low as possible and in the meantime reduce the variance of the inference time, since the model inference latency is affected by external and internal factors.  
\begin{itemize}
\item We analyze and profile the performance and characteristics of two-stage detection models, and find out that the first stage is the main contributor to the latency while the second stage significantly contributes to the large runtime variation due to the dynamic proposal numbers.
\item  Based on the analysis of the two-stage detection models, we propose \method, a systematic learning-based framework to achieve online thermal and latency variation management tailored for two-stage detectors. \method jointly adjusts the CPU and GPU frequency twice for each image inference with specialized design of the deep Q-network, experience replay buffer, and $\epsilon_t$-greedy cool-down action selection. 
\item We implement \method on different edge devices, including NVIDIA Jetson Orin Nano and Mi 11 Lite. The results indicate that \method achieves faster inference speed (by up to 30.8\% improvement) with reduced variation (by up to 72.8\%),  %and more stable inference (by up to 72.8\%), 
better thermal management, and a higher ratio of images meeting the latency constraint (by up to 43.8\%) than baselines under both static and dynamic environments.  % both under static environments, and adapt to dynamic environments.  
\end{itemize}

% Contribution:
% \todo{...}
% 1. Analyze the performance and characteristics of two-stage detection model in terms of latency, CPU GPU temperature, CPU GPU frequency on different platforms. 
% 2. systematic framework for fast and stable inference (mean, var, exact numbers here)
% 3. Previous works are not for two stage, why two stage is more challenging and close to real world. what is our specific design
% 4. formulate the online rl setting close to real world, evaluate the forgetting? online techniques? fast adapt

%% file: tex/2_related_work.tex
\section{Related Work}
% \subsection{\textbf{Existing DVFS techniques.}
% }
\textbf{Existing DVFS techniques.} 
DVFS is a common technique that dynamically adjusts the VF level of a processor for energy efficiency and thermal management. The VF scaling algorithm in a DVFS implementation, also known as governor, is typically provided by the processor manufacturer and controlled by the operating system. Examples of default governors in Linux are \texttt{ondemand} \cite{pallipadi2006ondemand} and \texttt{interactive} \cite{brodowski2013cpu}, which adjust CPU frequency based on predefined CPU utilization levels. Linux also employs a simplified version of \texttt{ondemand}, called \texttt{simple}$\_$\texttt{ondemand}, for GPU control.
While these governors aim to ensure stable performance and reduce power consumption, they have two limitations. First, they do not consider application performance. Second, having separate governors for CPU and GPU hampers efficient resource utilization within a limited power budget. Recently, a series of methods \cite{chuang2017adaptive,choi2019graphics} are proposed to design governors for various applications to tackle these limitations, yet the tasks are significantly different from deep learning (DL) tasks. Furthermore, most DVFS implementations still suffer from overheating problems. For devices like mobile phones, if the device temperatures reach a certain threshold, thermal throttling occurs, which significantly degrades application performance. zTT~\cite{kim2021ztt} takes both thermal throttling issues and DL-based applications into consideration. However, it suffers from a significant performance drop if directly applied to two-stage detectors. Latest work~\cite{lin2023workload} designs the governor from another perspective with % by considering 
multiple concurrent tasks, which is orthogonal to our design. 
% The performance of applications on edge devices is greatly influenced by the utilization of both the CPU and GPU, particularly their clock frequencies. In the initial phases of edge device platforms, the CPU takes charge of most operations, relegating the GPU to solely handling graphics-related tasks. Nonetheless, the advent of general-purpose GPUs (GPGPUs) revolutionized edge devices, empowering them to execute computationally intensive applications like deep learning (DL) with the aid of the GPU. Despite its vital role in optimizing performance, the appropriate coordination between CPU and GPU for a task still remains a problem for software developers.

% Object detection on edge devices, such as mobile phones and embedded systems, has gained significant attention in recent years due to the increasing demand for on-device processing.
% \subsection{Object Detection.}
\noindent \textbf{Object Detection.}  
There are two mainstreams of DNN-based object detection models, i.e., one-stage detectors and two-stage detectors. One-stage detectors perform object detection in a single pass without explicitly proposing regions. YOLO  \cite{redmon2016yolo} is one of the representative model that predicts bounding boxes and class probabilities directly from the full image. 
% Other one-stage detectors include SSD \cite{liu2016ssd}  that uses multiple default boxes to predict object locations and categories at different scales, and RetinaNet \cite{lin2017retinanet} that addresses the class imbalance problem in one-stage detectors using a focal loss. 
Though one-stage detectors are computationally efficient, they usually achieve lower accuracy compared to two-stage detectors.
Two-stage detectors propose region proposals before performing object classification and refinement. This approach has shown high accuracy in object detection and are more suitable to various application on edge devices that requires high precision such as perception in autonomous vehicles, suspicious activities detection in restricted areas, and environmental monitoring by drones. Faster R-CNN \cite{ren2015faster} introduces %region proposal networks (RPNs) 
a RPN to generate region proposals and a Fast R-CNN network for object classification and bounding box regression.
 Mask R-CNN \cite{he2017mask} extends Faster R-CNN to include an additional branch for instance segmentation, achieving state-of-the-art results in object detection and segmentation tasks.
 R-FCN~\cite{dai2016rfcn} proposes a region-based fully convolutional network, which enables efficient region-wise computation for object detection.

% \subsection{One-Stage Detectors}

% \begin{itemize}
%   \item YOLO (You Only Look Once): Introduced a real-time object detection system that 
%   \item SSD (Single Shot Multibox Detector) \cite{liu2016ssd}: Proposed a single-shot detector that uses multiple default boxes to predict object locations and categories at different scales.
%   \item RetinaNet \cite{lin2017retinanet}: Addressed the class imbalance problem in one-stage detectors using a focal loss, achieving better performance with dense object detection.
% \end{itemize}

%% file: tex/3_motivation.tex
\section{Motivations}

% \subsection{Thermal Issues}
\textbf{Thermal Issues.} 
With the advancement of edge devices, the capabilities of their processors have significantly increased, packing more power into compact spaces. However, this progress comes with the challenge of managing the heat generated during intensive computations. With the slim design of mobile phones and edge devices, thermal management becomes crucial to prevent severe performance degradation in such demanding scenarios. Multiple factors influence the temperature characteristics of these devices such as environmental temperature, on-device applications, thermal coupling among processors and battery. The mutual effects of these factors are too complicated to be characterized precisely, making the thermal management problem challenging.  

\input{tex/3.1_moti_latency}

% \subsection{Challenges of Two-Stage Detection Models}
\noindent \textbf{Challenges of Two-Stage Detection Models.}
Compared to one-stage detectors, % such as YOLO,   
two-stage detectors allow for more refined region proposals and improved object classification, resulting in enhanced detection performance \cite{ren2015faster,he2017mask,dai2016rfcn,lin2017feature,redmon2016yolo}, as shown in Fig. \ref{fig: model latency mean std} with a higher mAP. Specifically, they exhibit robustness in detecting objects of varying sizes with %the advantages brought by 
the region proposal stage to %that identifies 
identify potential object regions regardless of their size. This is especially beneficial in cases %where objects have significant size variations 
with objects of  significant size variations
in the same image, such as small objects in high resolution images obtained from aerial surveillance. 
%small objects detection such as in aerial surveillance, and detection tasks with high resolution case. 

Though enjoying these benefits, the practical deployment of two-stage detection models on edge devices faces the challenges of unstable inference.
To demonstrate the issue, we show the mean and variation of the inference latency for representative one-stage detector YOLOv5 \cite{ultralytics2021yolov5}, and two-stage detectors including FasterRCNN \cite{ren2015faster} and MaskRCNN \cite{he2017mask}  in Fig.~\ref{fig: model latency mean std}.
The mean and variation are calculated by the inference time on each dataset.
% On the KITTI dataset, FasterRCNN, MaskRCNN, and YOLOv5 exhibit inference speeds of $418.13\pm134.10$ms, $577.69\pm194.64$ms, and $78.80\pm2.75$ms, respectively.
On the KITTI dataset, FasterRCNN, MaskRCNN, and YOLOv5 exhibit inference variation of $134.10$ms, $194.64$ms, and $2.75$ms, respectively.
Similarly, on the VisDrone2019 dataset, the inference variation for these detectors are $198.85$ms, $200.18$ms, and $10.38$ms.
% Similarly, on the VisDrone2019 dataset, the inference speeds for these detectors are $648.88\pm198.85$ms, $673.59\pm200.18$ms, and $104.10\pm10.38$ms.
The latency variation for two-stage detectors %(FasterRCNN and MaskRCNN) 
is significantly greater than one-stage detectors. % on various datasets.
The unstable inference speed not only harms the user experience, but also wastes a lot of hardware resources \cite{gong2022all,gong2023condense}.
In applications involving object tracking, navigation, or control systems, large latency variations can negatively impact the accuracy of tracking and control algorithms.
Systems that integrate multiple sensors or modalities face challenges in maintaining synchronization when the latency of different sources varies significantly.
%This can lead to issues in aligning data from different sources, affecting the overall reliability of the system.
% \textcolor{red}{There need a figure showing the variation problem}.

The unstable latency problems for two-stage detectors are raised from two aspects. First, two-stage detection models are more \textbf{computationally intensive}, thus causing a higher risk of overheating. To mitigate the thermal issue, certain frequency adjustment is leveraged at the cost of varying inference time. Second, two-stage methods generate a \textbf{dynamic} number of \textbf{object proposals} for different image frames, thus incurring more uncertainty for the computation counts and thus inference speed. On the contrary, the feature maps in one-stage models are used to generate a static number of anchors or default boxes. 

% small objects detection

% drone

% scalability (high resolution)

% Online setting, why transfer setting is a simplified a version and not good enough

%% file: tex/3.1_moti_latency.tex
\begin{figure}
    \centering
    \includegraphics[width=.9\linewidth]{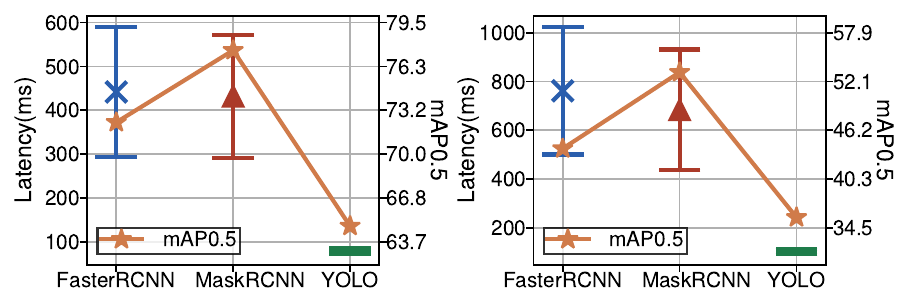}
       \vspace{-1em}
    \caption{\small The mean and variation of inference latency and precision for two-stage detectors (FasterRCNN, MaskRCNN) and one-stage detector (YOLOv5) on different datasets.} % (KITTI and Visdrone).}
    \label{fig: model latency mean std}
       \vspace{-1.5em}
    % \vspace{-7mm}
\end{figure}

% \begin{figure}
%     \centering
%         \subfloat[\centering KITTI changes]
%             {
%             \includegraphics[width=0.45\linewidth]{figs/motivation_variation_KITTI.pdf} 
%             \label{fig: temp var}
%             }
%         % \hfill
%         % \vspace{-1em}
%         \subfloat[\centering VisDrone2019]{
%             {
%             \includegraphics[width=0.45\linewidth]{figs/motivation_variation_VisDrone2019.pdf}
%             \label{fig: data var}
%             }
%         }%
%         \vspace{-1em}
%     \caption{
%     The mean and variation of inference latency and precision for two-stage detectors (FasterRCNN, MaskRCNN) and one-stage detector (YOLOv5) on different datasets (KITTI and Visdrone).
%     % The upper plot shows the temperature and the lower shows the latencies.
%     }
%     \label{fig: model latency mean std}
%     \vspace{-1em}
% \end{figure}

%% file: tex/4_method.tex
\section{\method Framework design}
%In this section, w
% We outline the approach of \method, starting by the problem formulation and then introducing the framework  with detailed design.
%We start by defining the optimization problem for two-stage detection models in Sec. \ref{sec:problem_formulation}, followed by an overview of the \method framework incorporating frequency scaling in Sec. \ref{sec:framework_overview}. Finally, we detail the specific DRL design in Sec. \ref{sec:drl_design}.

%\subsection{\method Framework Overview}

%We illustrate the framework overview of \method in Fig. \ref{fig:overview_framework}. \method is mainly composed of a DRL-based agent interacting with the environment and providing two actions for each image frame.   

% Online workflow or pipeline

% Frequency scaling for two-stage detection model

% Specialized DRL Agent Design for Two-Stage Detection Models

% Fast adaptation (online RL)
% e.g.buffer? fast adapt techniques? what else?

% system design? model compression?

%\subsection{Problem Formulation with Frequency Scaling for Two-Stage Detection Models}
\subsection{Problem Formulation} \label{sec:problem_formulation}

\subsubsection{Problem Formulation}

%As edge devices continue to advance in capabilities, their processors are becoming increasingly powerful and compact. However, this poses a challenge in managing the heat generated within such a confined space, particularly when engaging in intensive computations such as the execution of DNNs. 
It is necessary yet challenging to manage the heat, especially for the intensive DNN computations on edge devices.
At runtime, DVFS that adjust the CPU and GPU frequency ($f^{cpu}$ \& $f^{gpu}$) is an efficient method to decrease heat production from circuits and manage the inference speed dynamically. To provide better user experience %and achieve 
with more stable inference,  there are three requirements for the inference latency $l_i$ of the $i^{th}$ image: % frame: 
(i) $l_i$ should be as small as possible for fast responsiveness; (ii)  $l_i$ is better to meet the latency constraint $L$ %, which is 
posed by the applications, i.e., $l_i < L$; (iii) $l_i$ should maintain a small latency variation under various %dynamic running 
environments. 
%have to meet the deadline requirement $D$ and have a latency $l_i$ as small as possible, while maintaining a small latency variation under the dynamic running environment. 
Besides, to prevent the device from overheating that causes thermal throttling, the CPU temperature $T_i^{cpu}$ and GPU temperature $T_i^{gpu}$ should not exceed the pre-defined threshold value $T^{thres}$. 
%The latency $l_i$, CPU temperature $T_i^{cpu}$, and GPU temperature $T_i^{gpu}$ 
Thus, $l_i$, $T_i^{cpu}$, and $T_i^{gpu}$ are all highly influenced by %the running frequency 
$f^{cpu}$ and $f^{gpu}$. Scaling the frequency appropriately is a crucial problem to guarantee better user experience.  More specifically, the optimization problem can be formulated  mathematically as:
\begin{equation}
\small
    \begin{aligned}
         \min & \sum_{i=1}^I l_i + \alpha\cdot(l_i-\overline{l})^2 +\beta\cdot {U}(l_i-L) \\
        s.t. & \quad T_i^{cpu} \leq T^{thres}, \forall i \\
        & \quad T_i^{gpu} \leq T^{thres}, \forall i \\
    \end{aligned} \label{eq:optimization}
\end{equation}
where $\overline{l}$ represents the mean of the latency, ${U}$ is the step function with a value of 1 for inputs greater than 0,  %when the input is greater than 0, 
and 0 otherwise, $\alpha$ and $\beta$ are two hyperparameters adjusting their relative importance. % among the three terms.

\begin{figure}
    \centering
    \includegraphics[width=0.92 \linewidth]{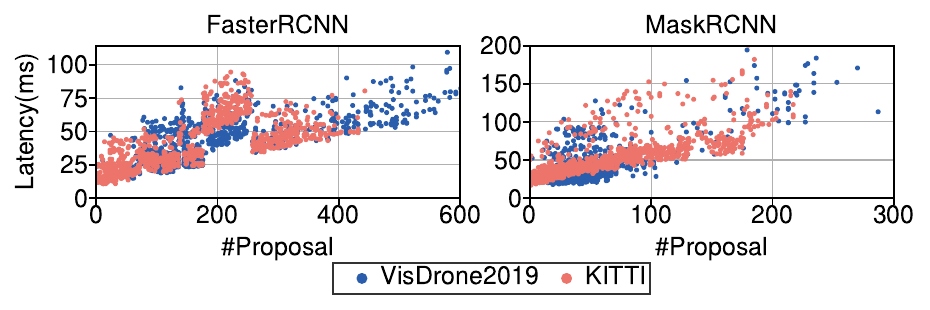}
       \vspace{-1em}
    \caption{\small Inference latency of the second stage for different numbers of proposals on FasterRCNN and MaskRCNN.}
    \vspace{-1em}
    \label{fig: num proposal vs latency}
\end{figure}

\subsubsection{Two-Stage Detection Models}
Understanding the object detection algorithms is the base to solve the optimization problem  in Eq.~(\ref{eq:optimization}).
Besides pre- %preocessing
and post-processing, the inference of a two-stage  detector   mainly has % composed of 
four parts. As shown in Figure \ref{fig:overview_framework},
it begins with a backbone such as ResNet-50 to extract the features.
Following the backbone is the RPN that  draws candidates of objects based on features to creates proposals. %in the image, 
The second stage %, the process 
begins with Region of Interest (RoI) pooling to create proposal feature maps with feature maps and region proposals.
%This step involves gathering feature maps and region proposals to create proposal feature maps.
The results %resulting proposals 
are then fed into the classifier, which utilizes fully connected layers to determine the object class of each proposal and establish a fixed bounding box.

The dynamic number of proposals generated in the first stage leads to dynamic computation counts in  following steps, and thus 
%causes the model to have 
more significant inference time variation compared to one-stage models with %that generate 
a static number of anchors/default boxes for detection. To verify this,
we demonstrate the correlation between the latency of the second stage and the number of proposals in Fig. \ref{fig: num proposal vs latency} by setting the CPU and GPU frequency at a fixed level. As  observed from the figure, on both datasets, a large number of proposals obtained by the first stage corresponds to a longer latency for the second stage.

\subsection{\method Framework with Frequency Scaling} \label{sec:framework_overview}

\begin{figure} [t]
     \centering
     \includegraphics[width=0.9\linewidth]{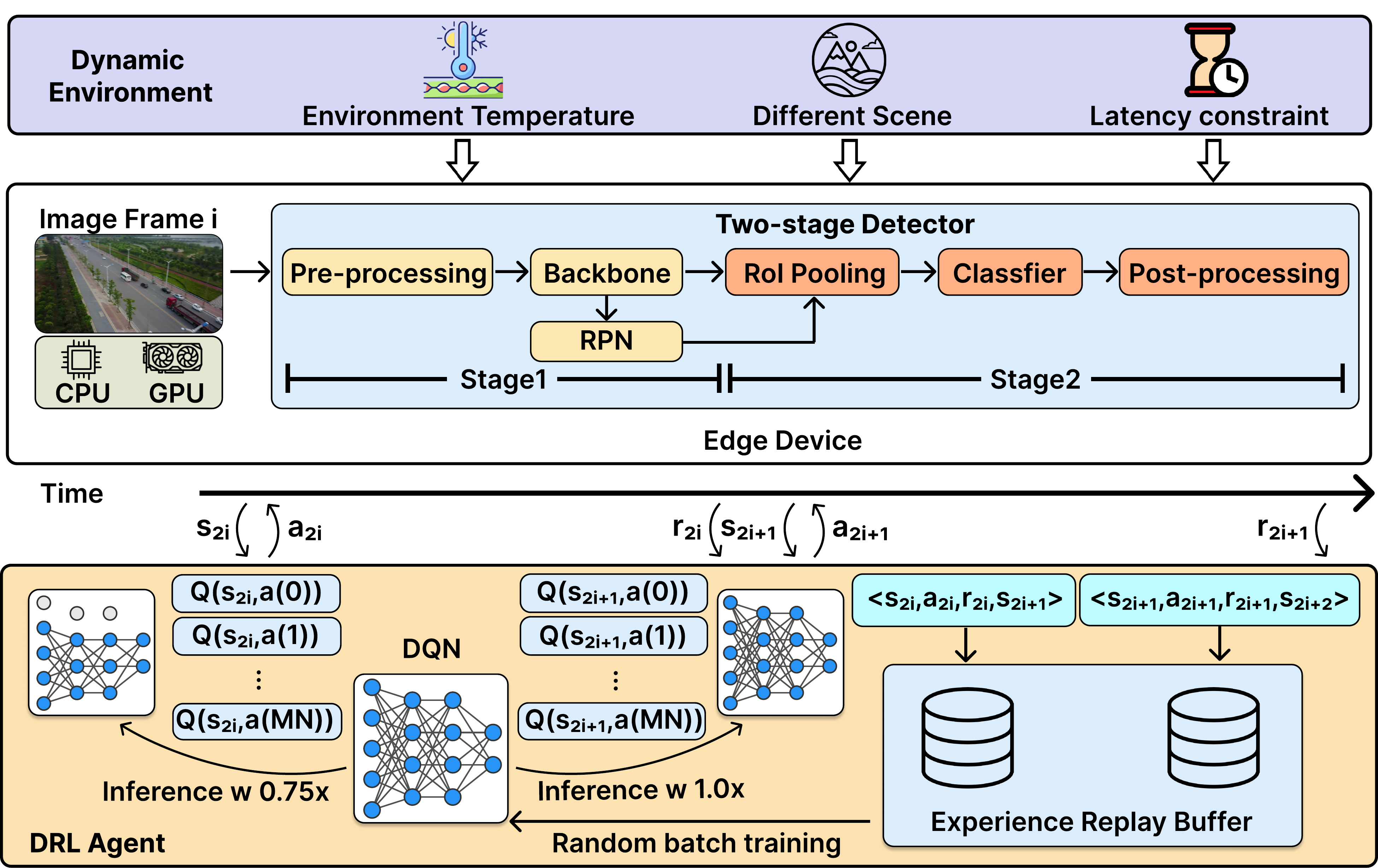}  
     % \vspace{-1em}
     \caption{\small Overview of \method. }
    \label{fig:overview_framework}  
    % \vspace{-1.5em}
    \vspace{-1.5em}
\end{figure}

%Based on the observation from the two-stage detectors, we can draw the following conclusions.
We make the following observations.
(i) As the backbone and RPN have a fixed amount of computations, the latency for these parts is influenced by the hardware execution environment (frequency).
(ii) The latency of the second stage is determined not only by the hardware environment, but also  the number of proposals from the first stage that varies across images.
To provide a stable inference speed for two-stage detectors, the frequency adjustment problem can be decomposed to \textit{when} and \textit{how} to adjust the frequency. 

% or periodically changing the frequency with fixed time interval \textcolor{red}{REF}
For the \emph{when} problem, we first perform a detailed profiling of the inference latency with fixed frequency. We observe that the  latency  of the first stage (including the preprocessing, backbone and RPN) %(i.e., from the start to the end of the first stage) 
takes about 80\% of the entire model latency, while the latency of the ROI, classifier, and post-processing %(from the start of the second stage to the end) 
takes about 20\%. Meanwhile, the runtime variation is mainly introduced by the second stage. E.g., for the FasterRCNN on KITTI, the runtime variation of its second stage can   be up to 159.5ms, where its entire inference is merely 441.9ms on average. We can conclude that with fixed frequency, the first stage is the main contribution for   latency while the second stage significantly contributes to the large runtime variation. 

Thus, for the  \emph{when} problem,  though scaling the frequency at the beginning of an image inference \cite{kim2021ztt} is the most standard method,  %these methods are 
it is not suitable for two-stage detectors, since it is not able to effectively reduce the latency variation from the second stage. 
%The chosen frequency for the first stage may not be suitable for the second stage, even at a fixed frequency level.
%Taking FasterRCNN on KITTI as an example, the runtime variation of its second stage can even be up to 159.5ms, where its entire inference is merely 441.9ms on average.
Besides, scaling the frequency at the start of the second stage solely is not desirable either, as it can hardly reduce the latency with the first stage as the main contributor.
%However, scaling the frequency after the first stage solely is not desirable either, as these two parts take 84.2\% and 15.8\% of the entire model inference.
Thus, based on the above discussion, the first \emph{when} question can be addressed by allowing two frequency decisions for each image inference, i.e.,  at the start of an image inference and after the creation of the proposals. 
%It can effectively minimize both the latency and the latency variation. 
%As the frequency decision is made twice for a single image, one may be curious about the overhead of frequency scaling.
Furthermore,  the overhead of frequency scaling is negligible with dozens of microseconds (compared to the DNN latency) on edge devices. 

%Typically, the frequency scaling on mobile devices only takes dozens of microseconds, which is negligible compared to the latency of a DNN.

% Thus, scaling the frequency once may not be enough to determine the frequency scaling appropriately for the two-stage object detection model. 

%Next it moves to the problem of \emph{how} to adjust the frequency. 
For the next \emph{how}  problem, directly solving the optimization problem as in Eq. (\ref{eq:optimization}) is infeasible. The frequency scaling problem is conducted in a dynamic and complex environment with varying images and CPU/GPU temperature.  The frequency to be taken not only depends on the current image and system information, but also the frequency decisions made in the past. Current frequency decision also affects the thermal situation in the near future. These features make our problem suitable to leverage DRL, where an agent learns how to make appropriate decisions through interaction with a dynamic environment. 
% \textcolor{red}{To deal with the second problem, considering the underlying difficulty in precisely modeling the environment change, such as transitions between temperatures with the configured CPU and GPU frequencies, and the sequential decision making process, DRL is well-suited to address the frequency scaling problem for two-stage detectors.  Next, we will elaborate our DRL design for \method.  }

% \subsubsection{Framework Overview}

Based on the above discussions, we illustrate the framework overview of \method in Fig. \ref{fig:overview_framework}. \method is mainly composed of a DRL-based agent interacting with the environment and providing two actions for each image frame.   
At the start of an image inference, it performs frequency scaling based on the current status. Next at the end of the first stage with number of proposals available, it  can further adjust the frequency for the optimization objectives. 

\subsection{DRL Design} \label{sec:drl_design}
% To reduce the runtime variation for the two-stage detection models, The GPU frequency is scaling twice for each image frame $i$.    
%The design of the DRL agent directly influences the solving of the problem. The design of the DRL agent further includes the definition of
We address the optimization of problem in Eq. (\ref{eq:optimization}) with a DRL approach, and  carefully design the state, action, and reward of the DRL agent in \method, as specified below. %We discuss the details below. 
\subsubsection{Action}
The DRL agent provides two separate actions $a_{2i}$ and $a_{2i+1}$ for the inference of the $i^{th}$ image frame. The first action $a_{2i}$ is given at the beginning of the $i^{th}$ image frame while the second action $a_{2i+1}$ is taken after the inference of the RPN. The two-action design is proposed to tackle the characteristics of the two-stage detection models as mentioned in Sec. \ref{sec:framework_overview}. The action set for both actions is the same. For a device equipped with $M$  CPU frequency levels and $N$ GPU frequency levels, the entire action space contains $M\times N$ different action choices. 
It is specifically defined as $\mathbf{a}=\{<f^{cpu,1}, f^{gpu,1}>, \cdots, <f^{cpu,m}, f^{gpu,n}>, \cdots, <f^{cpu,M}, f^{gpu,N}>\}$, where $<f^{cpu,m}, f^{gpu,n}>$ corresponds to scaling the CPU frequency to the $m^{th}$ level and GPU frequency to the $n^{th}$ level, respectively. 

\subsubsection{State}

Different from the action design that is the same for $a_{2i}$ and $a_{2i+1}$, the state differs for two consecutive time steps.
The state $s_{2i}$ observed at the beginning of the inference  of the $i^{th}$ image is formulated as a tuple with six elements represented as $\{S_{2i}, T_{2i}^{cpu}, T_{2i}^{gpu}, f_{2i}^{cpu}, f_{2i}^{gpu}, \Delta L_{2i} \}$. $S_{2i}$ indicates the current stage, $T_{2i}^{cpu}$ and  $T_{2i}^{gpu}$ represent the current CPU and GPU temperature, and $f_{2i}^{cpu}$ and $f_{2i}^{gpu}$ correspond to the current CPU and GPU clock frequency level.
$\Delta L_{2i}$ indicates the remaining time to meet the latency constraint. 
As for the state $s_{2i+1}$, it is formulated with a seven element tuple $\{S_{2i+1}, T_{2i+1}^{cpu}, T_{2i+1}^{gpu}, f_{2i+1}^{cpu}, f_{2i+1}^{gpu}, \Delta L_{2i+1}, P_{2i+1} \}$. The additional observation dimension $P_{2i+1}$ is the number of proposals for the $i^{th}$ image.
$P_{2i+1}$ is obtained based on the feature map from the backbone and the region proposals from the RPN, thus is only available at time step $2i+1$ for the $i^{th}$ image frame.

\subsubsection{Reward}
We propose a specialized reward design for \method.
To achieve the goal of stable inference for two-stage %detection models 
detectors,  %as in the optimization problem in Eq. (\ref{eq:optimization}), 
\method %have two objectives including optimizing
tries to optimize the latency while avoiding thermal throttling.
Specifically, the reward is defined as $r_{} = r^{time} + \lambda r^{temp}$, where $r^{time}$ stands for the reward for  latency  while $r^{temp}$ is the reward for  temperature.
%A penalty multiplier $p$ will be applied when each term is less than 0, respectively.
Specifically, $r^{time}$ is formulated as
% \begin{equation}
% \small
% \label{eq: time reward}
%     r^{time}_{i}(\Delta D_i) = \begin{cases}
%       \tanh({\Delta D_{i}}) + \frac{1}{1 + \sigma_{n}(\Delta D_{i})}, & if\ \Delta D_{i} > 0;\\
%       p \Delta D_{i}, & {otherwise,}
%       \end{cases}
% \end{equation}
\begin{equation}
% \small
\label{eq: time reward}
    r^{time}_{i} = \begin{cases}
      \tanh({\Delta L_{i}}) + \frac{1}{1 + \sigma_{n}(\Delta L_{i})}, & if\ \Delta L_{i} > 0;\\
      p \Delta L_{i}, & {otherwise,}
      \end{cases}
\end{equation}
where $\tanh{(\Delta L_i)} $ ensures fast inference and $\frac{1}{1 + \sigma_{n}(\Delta L_i)}$ constrains the latency variation, with $\sigma_{n}(\Delta L_i) $ representing the standard deviation calculated from the n most recent images. If $\Delta L_i < 0$, it means a violation of the time constraint, thus a penalty multiplier $p$ is applied. As for the temperature reward, it is defined as 
\begin{equation}
\centering
% \small
\label{eq: temp reward}
    r^{temp}_{i} = \begin{cases}
      1, & if\ T^{cpu} \leq T^{thres}\ and\  T^{gpu} \leq T^{thres};\\
      -p, & {otherwise,}
      \end{cases}
\end{equation}
where a positive reward is received as long as both $T^{cpu}$ and $T^{gpu}$ are below the device throttling temperature $T^{thres}$. However, when either temperature exceeds the throttling temperature $T^{thres}$, a penalty multiplier $p$ is applied.
% when the device temperature is lower than the throttling bound.
% Note that a penalty multiplier will be applied when the reward for each term is less than 0.

\subsubsection{Specialized DRL Agent Design for Two-Stage Detection Models}
% Due to the underlying difficulty in precisely modeling the environment change, such as transitions between temperatures with the configured CPU and GPU frequencies, 
\method resorts to a model-free DRL approach and leverages the popular Deep Q-Networks (DQN) algorithm \cite{mnih2015human}. The fundamental concept of Q-learning is based on the notion that if we had a function $Q^*$ capable of predicting the expected return when taking a specific action in a given state, it is easy to derive a policy that maximizes the rewards by simply following $\pi^*(s) = \arg\max_a Q^*(s, a)$. As DNNs can work as universal function approximators, the agent learns to approximate the optimal Q-value function $Q^*(s, a)$ with DNNs in the DQN algorithm. Different from other settings,  the observations of \method for state  $s_{2i}$ and $s_{2i+1}$ are different for the inference of the $i^{th}$ image. This indicates that there should be two sets of Q-value functions corresponding to the two different state-action pairs for each image. The most straightforward approach to tackle this is to leverage two separate DNNs (i.e., two Q-networks) for the approximation. But this \textbf{\underline{separation cuts off the correlations}} between the two actions for the inference of the same image frame. 

\textbf{To deal with this problem, we investigate whether we can keep \underline{only one DNN} for the Q-value approximation for these \underline{two sets} of state-action pairs.} We observe that the major difference between the first state-action pair $(s_{2i}. a_{2i})$ and second state-action pair $(s_{2i+1}. a_{2i+1})$  for the $i^{th}$ image frame is the presence of the proposal number $P_{2i+1}$, which can only be obtained after the inference of the first stage. Thus, we design the Q-network to be executed with two different widths (number of active channels)) $[\alpha\times, 1.0\times]$ configurations. The Q-value for the first state-action pair $(s_{2i}. a_{2i})$ is computed by only the first $\alpha\times$ (e.g., $0.75\times$ or 75\%) channels in each layer, while the Q-value of the second state-action pair $(s_{2i+1}. a_{2i+1})$ is obtained by executing the Q-network with full network width for each layer (i.e., $1\times$), as shown in Fig. \ref{fig:overview_framework}. With this design, the Q-value computations \textbf{\underline{share major parameters} \underline{and computations}}. Two separate experience replay buffers are leveraged to store the observed transitions, i.e., one is to store transitions $<s_{2i}, a_{2i}, r_{2i}, s_{2i+1}>$ at $2i$ time step while the other is to store $<s_{2i+1}, a_{2i+1}, r_{2i+1}, s_{2i+2}>$ for $2i+1$. During training, a batch of random samples is selected in the corresponding experience replay buffer to update the Q-network. In particular, at time step $2i$, the sampled transitions are used to update the Q-network with $\alpha\times$ width, while the remaining weights are not updated. 

% \subsubsection{Fast Adaptation to Environment Change}
% During training, DQN uses an experience replay buffer to store the agent's experiences (state, action, reward, next state, done) while interacting with the environment. The agent samples batches of experiences from the replay buffer to compute the

\subsubsection{$\epsilon_t$-greedy cool-down action selection}
\label{sec: greedy cd}
Thermal issues are inevitable during the DRL training and inference phases.
Unlike larger devices~(such as servers and desktops) with fans or powerful cooling systems, heat can not be effectively dissipated on edge devices.
The ineffective heat dissipation leads to delayed responsiveness for DRL actions, yielding difficulty in the DQN algorithm with $\epsilon$-greedy that explores a random action with probability $\epsilon$.
To avoid choosing higher frequency when the device is already overheated ($T^{cpu}$ or $T^{gpu}$ is higher than $T^{thres}$), zTT \cite{kim2021ztt} introduces a \textit{cool-down action} that specifically selects a random CPU and GPU frequency which is lower than the current status in such circumstances.
However, this random action prevents the agent from learning reasonable action selections when the temperature is high, thus the agent is not able to provide good action selection in such cases.
To avoid this issue, \method introduces $\epsilon_{t}$-greedy cool-down action selection, where $\epsilon_{t}$ is initialized between $[0, 1]$. When the device is overheated, \method selects a random frequency lower than the current setting by the probability $\epsilon_{t}$.
Otherwise, the action is selected according to the output of the Q-network.
Each time the cool-down action is triggered, the probability $\epsilon_{t}$ decays sinusoidally as the agent accumulates more experience in handling the overheating case. 
Implementing $\epsilon_{t}$-greedy cool-down action selection prevents severe performance drops due to thermal issues in the early training phase, promoting smoother convergence of the Q-network. Simultaneously, it enables the agent to gradually improve its ability to manage overheating scenarios throughout the training process.

% To avoid 
% However, the cool-down action must be applied during both the training and inference phases, introducing additional overhead other than Q-network.
% We benchmark the inference of Q-network and \textit{cool-down action} on the server, where Q-network only takes 0.42ms while \textit{cool-down action} consumes an additional 0.11ms.
% Optimizing the \textit{cool-down action} can speed up the Q-network by 20.75\%.
% To address this dilemma, we proposed the $\epsilon_{t}$-greedy cooldown action selection.
% Similar to the $\epsilon$-greedy, an hyper-parameter $\epsilon_{t}$ is initialized between $[0, 1]$.
% The cooldown action as zTT is selected by the probability $\epsilon_{t}$.
% Otherwise, the action is selected according to the output of the Q-network.
% The $\epsilon_{t}$ decays when the client device is overheated.
% We highlight that the $\epsilon_{t}$-greedy cooldown action selection is only applied during the training phase and will not cause overhead during inference.

\input{tex/5.1_exp_stable}

\subsection{\method Implementation}
\method is implemented in Python for two edge devices. 
\begin{enumerate*}[label=(\roman*)]
    \item \label{dev: jetson}
    NVIDIA Jetson Orin Nano  with a 6-core Arm Cortex-A78AE v8.2 64-bit CPU~(1.5GHz), 1024-core NVIDIA Ampere architecture GPU with 32 Tensor Cores~(625MHz), and an 8GB 128-bit LPDDR5 memory.
    \item \label{dev: mobi}
    Mi 11 Lite equipped Qualcomm SnapDragon 780G chipset with a Qualcomm Kryo 670 Octa-core CPU (1$\times$2.40 GHz Cortex-A78 \& 3$\times$2.22 GHz Cortex-A78 \& 4$\times$1.90 GHz Cortex-A55) and a Qualcomm Adreno 642 GPU.
\end{enumerate*}
Fig.~\ref{fig:overview_framework} illustrates an overview of \method implementation.
The \method agent is executed with an NVIDIA 2080Ti GPU and controls the CPU and GPU frequency of client devices.
% We use Pytorch on~(\ref{dev: jetson}) and MNN~\cite{MNN} on~(\ref{dev: mobi}) to enable DNN inference task on edge devices. 
Data between the client~(edge device) and agent is transmitted through the \texttt{socket}.
Note that data can also be transmitted via wired or wireless networks to adapt to various scenarios.
The observation of each state is directly collected through the \texttt{sysfs} in the Linux kernel and Android kernel for \ref{dev: jetson} and \ref{dev: mobi}, respectively.

\subsubsection{\method Training}
The Q-network architecture is designed as a 4-layer MLP that works at two widths [0.75$\times$, 1$\times$].
% Each layer is followed by the ReLU activation and the slimable batch normalization.
The Q-network in \method agent selects actions based on typical $\epsilon$-greedy and $\epsilon_{t}$-greedy cool-down action selection proposed in Sec.~\ref{sec: greedy cd}.
The agent collects observation samples, calculates rewards, and stores samples in the two sets of experience replay buffers.
Then, the Q-network is trained for 10,000 iterations using Adam optimizer with $\beta_1=0.9, \beta_2=0.99$, and the learning rate is set to 0.01 with cosine decay.

\subsubsection{Overhead analysis}
The overhead incurred by the \method is trivial.
It comes from executing the Q-network of the agent and data transmission between the agent and the client device.
The Q-network latency is benchmarked on a desktop with an NVIDIA GeForce 2080Ti GPU, which only consumes 0.42ms on average.
The data transmission via \texttt{socket} takes 1.92ms/message.
Thus, the overall overhead of \method is 8.52ms/inference, which is marginal compared to the inference latency of the two-stage detection model.

%% file: tex/5.1_exp_stable.tex
\begin{figure}[t]
    \centering
        \subfloat[\centering VisDrone2019 dataset]
            {\includegraphics[width=0.47\linewidth]{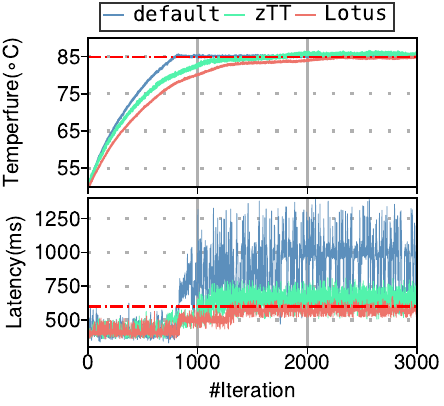} }
        % \hfill
        % \vspace{-1em}
        \subfloat[\centering KITTI dataset]{
            {\includegraphics[width=0.47\linewidth]{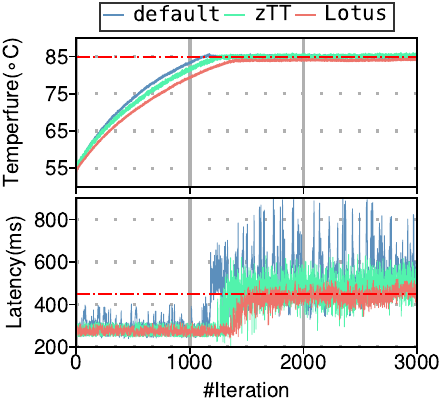} }
        }%
        \vspace{-1.em}
    \caption{ \small
    Comparison on Jetson Orin Nano with FasterRCNN. Red dashed lines indicate the throttling bound and latency constraint.
    % The upper plot shows the temperature and the lower shows the latencies.
    }
    \label{fig: exp faster_rcnn_static}
    \vspace{-1em}
\end{figure}
\begin{figure}[t]
    \centering
        \subfloat[\centering VisDrone2019 dataset]
            {\includegraphics[width=0.47\linewidth]{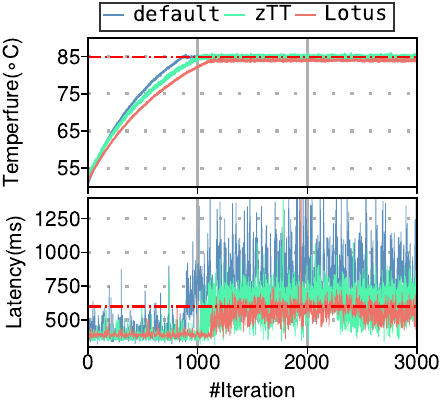} }
        % \hfill
        % \vspace{-1em}
        \subfloat[\centering KITTI dataset]{
            {\includegraphics[width=0.47\linewidth]{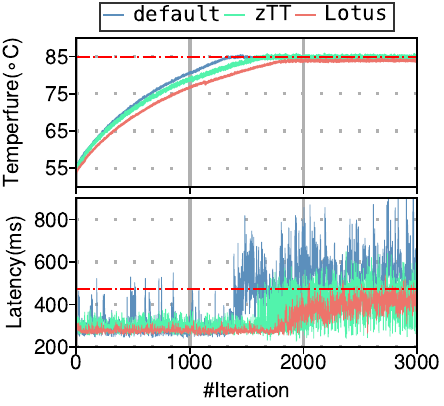} }
        }%
        \vspace{-1.em}
    \caption{
    \small Evaluation on Jetson Orin Nano with MaskRCNN. Red dashed lines indicate the throttling bound and latency constraint.
    }
    \label{fig: exp mask_rcnn_static}
    \vspace{-1em}
\end{figure}

%% file: tex/5_experiments.tex
\section{Experiments}

%%%%%%%%% Figures %%%%%%%%%% 
%%%%%%%%% Figures %%%%%%%%%% 

% \textcolor{red}{1. static environment results [latency\&temp curve for diff dataset [KITTI, visdrone], model [fasterrcnn, maskrcnn, do we have others?], and temp, platform(jetson board\&mobile); compare with default\&ztt]}

% \textcolor{red}{2. dynamic envrioment results [when temp/dataset changes, how the latency\&temp\&reward looks like, similar to Fig. 18 in ztt paper]}

% \textcolor{red}{3. time consumption of our approach}  

We implement \method on Jetson Orin Nano and Mi 11 Lite to demonstrate its 
advantages of maintaining stable DNN inference while preventing overheating.
% and maintaining stable inference.
\input{tabs/static}

\subsection{Experiment Setup}

\subsubsection{Baselines.}
We compare with the default governors and the latest learning-based thermal management governor zTT \cite{kim2021ztt}. % on NVIDIA Jetson Orin Nano and  Mi 11 Lite. 
The default governors for the CPU and GPU on NVIDIA Jetson Orin Nano are \texttt{schedutil} and \texttt{nvhost\_podgov}, respectively.  Mi 11 Lite adopts \texttt{schedutil} and \texttt{msm-adreno-tz} as its governor to control its CPU and GPU frequency. The state-of-the-art zTT \cite{kim2021ztt} scales CPU and GPU frequency jointly with a DRL approach. %, which is considered as the state-of-the-art.
% \subsubsection{State-of-the-Art Works}
% The \texttt{schedutil} and \texttt{nvhost\_podgov} governors are used for CPU and GPU on NVIDIA Jetson Orin Nano, respectively, which are the default governors. 
% These governors dynamically adapt the CPU and GPU frequency according to the application workloads.
% Similarly, Mi 11 Lite adopts \texttt{schedutil} and \texttt{msm-adreno-tz} as its governor to control its CPU and GPU frequency.
% When the device temperature surpasses the throttling threshold, these governors will adjust the CPU and GPU frequency to mitigate the overheating problem.
% Here, we denote these governor-based managements as \texttt{default}.

% zTT~\cite{kim2021ztt} is a recent learning-based thermal optimization method.
% It aims to optimize the CPU and GPU frequency via DVFS of the target device to mitigate the thermal issue while preventing application performance degradation.
% zTT is the current state-of-the-art in learning-based DVFS with thermal optimization. 
% We carefully adapt the zTT framework to two-stage detection model scenarios and our evaluation devices.
% \todo{brief intro to zTT}

\subsubsection{Evaluation Environments}
\label{sec: eval envs}
%To verify the benefits of \method, 
We evaluate \method %and compare it 
in comparison with baselines across diverse settings that simulate real-world usage environments.
We test on the KITTI  and VisDrone2019 datasets, %are selected as they reflect the uses of object detection 
which are commonly used in autonomous vehicles and drones. 
The FasterRCNN and MaskRCNN are selected as the on-device two-stage detectors. %object detection models for running on the device.
These two well-known models are %famous models in this field and 
the basis of succeeding two-stage detectors. % object detection models.
Note that different latency constraints in \method are applied for different datasets and models due to their varied computation demands.
We also test in both static  and dynamic  external environments for practical considerations. % that simulate the changes in the external surroundings.

\subsection{Experiment Results}

\subsubsection{\method on Static External Environments}
%To demonstrate the advantages of the \method under a static environment, we evaluate it on various detectors, datasets, and devices as mentioned in Sec.\ref{sec: eval envs}.
The static external environment is defined by the normal $25^\circ$C indoor environment without external cooling systems.
%of $25^\circ$C and no external cooling system applied.
We compare the device temperature (averaged between the CPU temperature and GPU temperature) and detector latency %to the \texttt{default} and zTT 
by executing the detector 3,000 iterations on the device (each iteration corresponds to processing one image).
For NVIDIA Jetson Orin Nano, the results are shown in Fig. \ref{fig: exp faster_rcnn_static} for FasterRCNN, Fig. \ref{fig: exp mask_rcnn_static} for MaskRCNN, and Tab. \ref{table: exp static results}. %From the results 
We can observe that \method consistently achieves better performance by maintaining a lower device temperature, faster average inference speed, smaller latency variation, and higher ratio of meeting preset time constraints (i.e., higher satisfaction rate). Specifically, %in Fig. \ref{fig: exp faster_rcnn_static}, throughout the entire inference process, \method performs more stable compared to baseline methods with a faster inference speed and smaller variation. Take 
for MaskRCNN on VisDrone2019, % as an example,  
\method reduces the latency by 30.8\% compared to the default and 9.1\% compared to zTT, with a lower device temperature consistently. Meanwhile, the \method latency variation measured by standard deviation is reduced by 72.8\% and 38.1\% %with \method 
compared to the two baselines. Furthermore, \method improves the satisfaction rate %ratio of images meeting the time constraint 
by  35.9\% and 24.8\% compared to the two baselines.

% \todo{Fig.~\ref{fig: exp faster_rcnn_static}, Fig.~\ref{fig: exp mask_rcnn_static}, Tab.~\ref{table: exp static results} analysis.}
%\input{tabs/static}

% The detector latency is evaluated as the average latency and the standard deviation of latency.
%We further show the performance of \method on mobile devices by implementing it on  Mi 11 Lite, and 
The results on Mi 11 Lite are shown in Fig. \ref{fig: exp mobi faster} and Tab. \ref{table: exp mobi results}. Similar to %the observations obtained on 
Jetson Orin Nano, \method achieves better performance than all baselines across all measures. For instance, \method decreases the latency variation by 29.4\% and 9.5\% for MaskRCNN on KITTI. The ratio of images meeting time constraints are improved to 92.5\% by \method for FasterRCNN on VisDrone2019. %Combining the results on both platforms 
Based on the above results, %we can see that 
\method is effective in thermal and latency variation management.
\input{tex/5.3_exp_mobi}
\input{tabs/mobi}

\subsubsection{\method on Dynamic External Environments}

%When deployed on edge devices, \method should process the ability to handle with dynamic environment change.  
The external environment of the devices may change dynamically. 
Thus, to exhibit the robustness of \method across various environments, we test the edge device with diverse environmental changes. 

\paragraph{Temperature changes}
% \todo{explain why temperature changes}
%Due to the exposure to external environments,  
An edge device typically works under varying temperatures. For instance, a mobile device %carried by a person might 
may switch between the warm indoor and the cold outdoor frequently in the winter. A drone operating in open airspace can experience very different outside temperatures due to the altitude and weather conditions.
The change of environmental temperature challenges the robustness of \method.
To verify its robustness against temperature fluctuating, we define two temperature zones: 
\begin{enumerate*}[label=(\roman*)]
    \item warm zone~($25^\circ$C)
    \item cold zone~($0^\circ$C).
\end{enumerate*}
The device is placed at different temperature zones during the inference.
Fig.~\ref{fig: temp var} shows the results of changing temperature zones when using the MaskRCNN detector on  VisDrone2019. %From the results 
We can see that \method can smoothly and quickly adapt to the temperature changes %by providing a more stable device temperature and latency than baselines. 
with consistently lower temperature, faster inference, smaller latency variation, and higher satisfaction rate. With lower outside temperature and better cooling conditions, \method seeks to keep low latency and variations. With higher temperature, \method tends to prevent overheating while meeting the latency constraint. %When the outside temperature becomes lower, the thermal headroom becomes more generous, thus \method seeks to improve the inference speed. On the contrary, when the outside temperature becomes higher, which corresponds to a more insufficient thermal bedroom, \method tend to reduce the running frequency to prevent overheating.    

\input{tex/5.2_exp_dynamic}
\paragraph{Domain changes}
It is common for a device to carry multiple tasks on different domains.
For example, the search and rescue drone is expected to detect vehicles on the street and identify missing persons in the wild.
Furthermore, the domain change is usually accompanied with different time constraint settings as different tasks usually have different requirements.
Here, we switch the dataset from KITTI to VisDrone2019 during inference, as shown in Fig.~\ref{fig: data var}. The results indicate that \method performs better than baselines when the task domain switches with a more stable inference and thermal management. %Combining the results 
We can conclude that \method can better adapt to environmental changes and utilize hardware resources. % more efficiently. 
% \todo{deadline also changes, analysis}

% \subsubsection{\method on Mobile Device}

\section{Conclusion}
This paper proposes \method, a framework that tackles the challenges of unstable inference and thermal issues on edge devices for two-stage detectors by dynamically scaling CPU and GPU frequencies via DRL.  Experiments demonstrate consistent improvements in latency and temperature control, making \method a promising solution for stable and efficient edge-based applications.
% This paper proposes \method, a framework that tackles the challenges of unstable inference and thermal issues on edge devices for two-stage detectors. By dynamically scaling CPU and GPU frequencies via DRL, \method significantly reduces latency variations.  Experiments demonstrate consistent improvements in inference speed and temperature control, making \method a promising solution for stable and efficient edge-based applications.

%  By dynamically adjusting CPU and GPU frequencies using DRL, \method significantly reduces latency variations.

%% file: tabs/static.tex
\newcommand{\shadow}[1]{\cellcolor[HTML]{EFEFEF}}
\begin{table}

\small
\caption{ \small Quantitative results on 
% FasterRCNN and MaskRCNN detector with KITTI and VisDrone2019 datasets at 
Jetson Orin Nano.
The $\overline{l}$ and $\sigma_{l}$ indicate the mean and standard deviation of latency.
$R_{L}$ indicates the ratio of meeting the preset latency constraint (satisfaction rate).}
% \left
\resizebox{0.9\columnwidth}{!}{
% \footnotesize

% \usepackage{multirow}
% \renewcommand{\arraystretch}{1.1}
\renewcommand\tabcolsep{2pt}
% \scalebox{1}
% {

\begin{tabular}{c||cc}
\toprule

% \begin{tabular}{c}
% \multirow{2}{*}{\textbf{Model}} \\
% \\
% \hline
% \hline

% % \midrule
% \multirow{3}{*}{{\scriptsize{\rotatebox[origin=c]{0}{FasterRCNN}}}}
% \\ % row1
% \\ % row2
% \\ % row3
% % FasterRCNN \\
% % FasterRCNN \\
% % FasterRCNN \\
% \hline
% \multirow{3}{*}{{\scriptsize{\rotatebox[origin=c]{0}{MaskRCNN}}}}
% \\ % row1
% \\ % row2
% \\ % row3
% % MaskRCNN \\
% % MaskRCNN \\
% % MaskRCNN \\
% \end{tabular}
% &

% \renewcommand\tabcolsep{1.5pt}
\begin{tabular}{c||c}
\multirow{2}{*}{\textbf{Detector}} & \multirow{2}{*}{\textbf{Method}} \\
\\
\hline
\hline
% \midrule
\multirow{3}{*}{{{\rotatebox[origin=c]{0}{FasterRCNN}}}}
& default \\
& zTT \\
& \cellcolor[gray]{.9} \textbf{\method} \\
% \midrule
\hline
\multirow{3}{*}{{{\rotatebox[origin=c]{0}{MaskRCNN}}}}
& default \\
& zTT \\
& \cellcolor[gray]{.9} \textbf{\method} \\
\end{tabular}
&
% \tabcolsep{1.5pt}
% \renewcommand\tabcolsep{1.5pt}
\begin{tabular}{ccc}
 \multicolumn{3}{c}{\textbf{KITTI}} \\
% \midrule
{$\overline{l}$}{(ms)}$\downarrow$ & {$\sigma_{l}$}{(ms)}$\downarrow$ &  $R_{L}\uparrow$ \\
\hline
\hline
% \midrule
434.6 & 139.8    &  51.4\%    \\
363.7 & 85.6     &  55.5\%   \\
\cellcolor[gray]{.9}\textbf{343.2} & \cellcolor[gray]{.9}\textbf{68.6}  &  \cellcolor[gray]{.9}\textbf{66.5\%}       \\
% \midrule
\hline
443.9 & 148.0     &  59.8\%   \\
408.3 & 111.7      &  87.1\%  \\
\cellcolor[gray]{.9}\textbf{388.5} & \cellcolor[gray]{.9}\textbf{88.9}  & \cellcolor[gray]{.9}\textbf{95.2\%}      \\
\end{tabular}
&
\begin{tabular}{ccc}
\multicolumn{3}{c}{\textbf{VisDrone2019}} \\
% \midrule
{$\overline{l}$}{(ms)}$\downarrow$ & {$\sigma_{l}$}{(ms)}$\downarrow$ &  $R_{L}\uparrow$ \\

\hline
\hline
% \midrule
686.0       & 241.1    &  29.4\%    \\
577.6        & 167.5    &  46.3\%   \\
\cellcolor[gray]{.9}\textbf{523.5}       & \cellcolor[gray]{.9}\textbf{102.9} & \cellcolor[gray]{.9}\textbf{71.1\%} \\
% \midrule
\hline
768.4       & 260.4   &  39.0\%     \\
584.3       & 114.2    &  50.1\%    \\
\cellcolor[gray]{.9}\textbf{531.4}       & \cellcolor[gray]{.9}\textbf{70.7}       &  \cellcolor[gray]{.9}\textbf{74.9\%}    \\                                        
 \end{tabular}
\\

\bottomrule

\end{tabular}
}
 \vspace{-1em}
\label{table: exp static results}
\end{table}

%% file: tex/5.3_exp_mobi.tex
\begin{figure}
    \centering
        \subfloat[\centering VisDrone2019 dataset]
            {
            \includegraphics[width=0.46\linewidth]{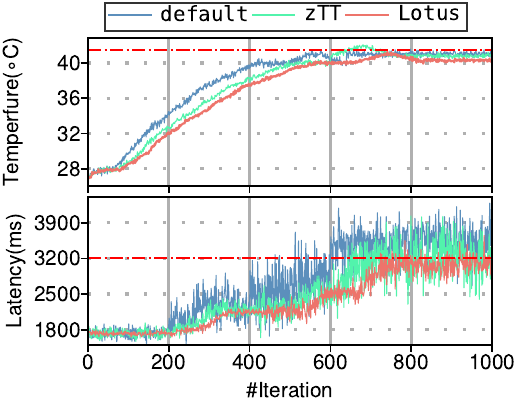} 
            \label{fig: mobi vsdrone}
            }
        \subfloat[\centering KITTI dataset]{
            {
            \includegraphics[width=0.46\linewidth]{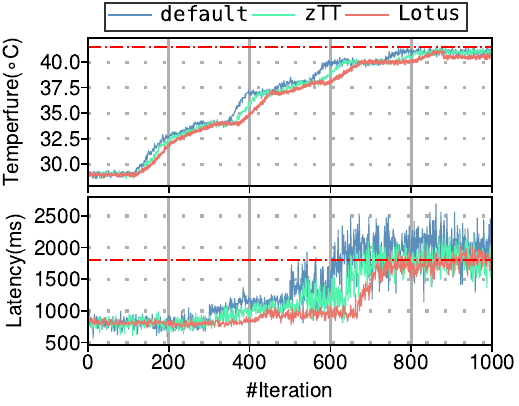}
            \label{fig: mobi vitti}
            }
        }%
        \vspace{-1em}
    \caption{
   \small Evaluations on  Mi 11 Lite with FasterRCNN. Red dashed lines indicate the throttling bound and latency constraint.
    % The upper plot shows the temperature and the lower shows the latencies.
    }
    \label{fig: exp mobi faster}
    \vspace{-0.5 em}
\end{figure}

%% file: tabs/mobi.tex
\begin{table}
\small
\caption{\small Quantitative results on 
% FasterRCNN and MaskRCNN detector with KITTI and VisDrone2019 datasets at
Mi 11 Lite 5G.
The $\overline{l}$ and $\sigma_{l}$ indicate the mean and standard deviation of latency.
$R_{L}$ indicates the ratio of meeting the preset latency constraint (satisfaction rate).}
% \left
\resizebox{0.9\columnwidth}{!}{
% \footnotesize

% \usepackage{multirow}
% \renewcommand{\arraystretch}{1.1}
\renewcommand\tabcolsep{2pt}
% \scalebox{1}
% {

\begin{tabular}{c||cc}
\toprule

% \begin{tabular}{c}
% \multirow{2}{*}{\textbf{Model}} \\
% \\
% \hline
% \hline

% % \midrule
% \multirow{3}{*}{{\scriptsize{\rotatebox[origin=c]{0}{FasterRCNN}}}}
% \\ % row1
% \\ % row2
% \\ % row3
% % FasterRCNN \\
% % FasterRCNN \\
% % FasterRCNN \\
% \hline
% \multirow{3}{*}{{\scriptsize{\rotatebox[origin=c]{0}{MaskRCNN}}}}
% \\ % row1
% \\ % row2
% \\ % row3
% % MaskRCNN \\
% % MaskRCNN \\
% % MaskRCNN \\
% \end{tabular}
% &

% \renewcommand\tabcolsep{1.5pt}
\begin{tabular}{c||c}
\multirow{2}{*}{\textbf{Detector}} & \multirow{2}{*}{\textbf{Method}} \\
\\
\hline
\hline
% \midrule
\multirow{3}{*}{{{\rotatebox[origin=c]{0}{FasterRCNN}}}}
& default \\
& zTT \\
& \cellcolor[gray]{.9} \textbf{\method} \\
% \midrule
\hline
\multirow{3}{*}{{{\rotatebox[origin=c]{0}{MaskRCNN}}}}
& default \\
& zTT \\
& \cellcolor[gray]{.9} \textbf{\method} \\
\end{tabular}
&
% \tabcolsep{1.5pt}
% \renewcommand\tabcolsep{1.5pt}
\begin{tabular}{ccc}
 \multicolumn{3}{c}{\textbf{KITTI}} \\
% \midrule
{$\overline{l}$}{(ms)}$\downarrow$ & {$\sigma_{l}$}{(ms)}$\downarrow$ &  $R_{L}\uparrow$ \\
\hline
\hline
% \midrule
1377.5 & 525.1    &  70.9\%    \\
1260.9 & 448.2     &  83.3\%   \\
\cellcolor[gray]{.9}\textbf{1185.8} & \cellcolor[gray]{.9}\textbf{429.9}  &  \cellcolor[gray]{.9}\textbf{89.7\%}       \\
% \midrule
\hline
1652.1 & 781.8     &  61.3\%   \\
1582.7 & 610.5      &  79.8\%  \\
\cellcolor[gray]{.9}\textbf{1429.5} & \cellcolor[gray]{.9}\textbf{552.3}  & \cellcolor[gray]{.9}\textbf{91.5\%}      \\
\end{tabular}
&
\begin{tabular}{ccc}
\multicolumn{3}{c}{\textbf{VisDrone2019}} \\
% \midrule
{$\overline{l}$}{(ms)}$\downarrow$ & {$\sigma_{l}$}{(ms)}$\downarrow$ &  $R_{L}\uparrow$ \\

\hline
\hline
% \midrule
2728.0       & 761.5    &  63.3\%    \\
2509.7        & 649.3    &  79.7\%   \\
\cellcolor[gray]{.9}\textbf{2421.0}       & \cellcolor[gray]{.9}\textbf{558.7} & \cellcolor[gray]{.9}\textbf{92.5\%} \\
% \midrule
\hline
3241.9       & 725.5   &  40.1\%     \\
2972.5       & 621.7    &  59.4\%    \\
\cellcolor[gray]{.9}\textbf{2649.5}       & \cellcolor[gray]{.9}\textbf{591.2}       &  \cellcolor[gray]{.9}\textbf{83.8\%}    \\                                        
 \end{tabular}
\\

\bottomrule

\end{tabular}
}

 \vspace{-1em}
\label{table: exp mobi results}
\end{table}

%% file: tex/5.2_exp_dynamic.tex
\begin{figure}
    \centering
        \subfloat[\centering Temperature changes]
            {
            \includegraphics[width=0.46\linewidth]{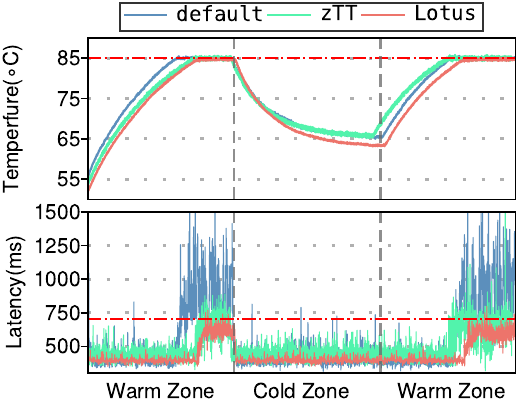} 
            \label{fig: temp var}
            }
        % \hfill
        % \vspace{-1em}
        \subfloat[\centering Domain changes]{
            {
            \includegraphics[width=0.46\linewidth]{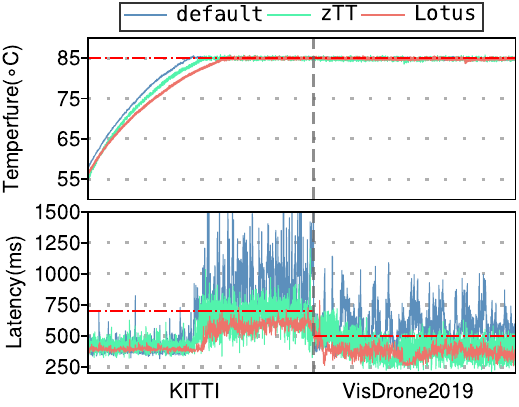}
            \label{fig: data var}
            }
        }%
        \vspace{-1em}
    \caption{ \small
    Performance when the environment changes. Red dashed lines indicate the throttling bound and latency constraint.
    % The upper plot shows the temperature and the lower shows the latencies.
    }
    \label{fig: exp mask dynamic}
    \vspace{-1.5 em}
\end{figure}